\documentclass[letterpaper, 10 pt, conference]{ieeeconf}  

\IEEEoverridecommandlockouts                              

\makeatletter
\def\endthebibliography{%
	\def\@noitemerr{\@latex@warning{Empty `thebibliography' environment}}%
	\endlist
}
\makeatother                                    

\makeatletter
\newcommand*\titleheader[1]{\gdef\@titleheader{#1}}
\AtBeginDocument{%
  \let\st@red@title\@title
  \def\@title{%
    \bgroup\normalfont\large\centering\@titleheader\par\egroup
    \vskip0.5em\st@red@title}
}
\makeatother

\usepackage[noadjust]{cite}
\usepackage{amsmath,amssymb,amsfonts}
\usepackage{algorithmic}
\usepackage[pdftex]{graphicx}
\usepackage{textcomp}
\usepackage{xcolor}
\usepackage{subfigure}
\usepackage{nccmath}
\usepackage{balance}
\usepackage{multirow}
\usepackage{hyperref}
\usepackage{mathtools}

\title{\LARGE \bf
A Deep Learning-based Global and Segmentation-based Semantic Feature Fusion Approach for Indoor Scene Classification
}

\titleheader{Published at Pattern Recognition Letters 2024 (DOI: 10.1016/j.patrec.2024.01.022)} 

\author{Ricardo Pereira$^{1}$, Tiago Barros$^{1}$, Lu\'{i}s Garrote$^{1}$, Ana Lopes$^{1,2}$, Urbano J. Nunes$^{1}$ 
\thanks{$^{1}$Authors are with the University of Coimbra, Institute of Systems and Robotics, Department of Electrical and Computer Engineering, Portugal. Emails: \footnotesize\{ricardo.pereira,~tiagobarros,~garrote,~anacris,~urbano\}@isr.uc.pt.}
\thanks{$^{2}$Author is also with the Polytechnic Institute of Tomar, Portugal.}
}

\begin{document}

\maketitle
\thispagestyle{empty}
\pagestyle{empty}

\begin{abstract}
This work proposes a novel approach that uses a semantic segmentation mask to obtain a 2D spatial layout of the segmentation-categories across the scene, designated by segmentation-based semantic features (SSFs). These features represent, per segmentation-category, the pixel count, as well as the 2D average position and respective standard deviation values. Moreover, a two-branch network, GS$^2$F$^2$App, that exploits CNN-based global features extracted from RGB images and the segmentation-based features extracted from the proposed SSFs, is also proposed.
GS$^2$F$^2$App was evaluated in two indoor scene benchmark datasets: the SUN RGB-D and the NYU Depth V2, achieving state-of-the-art results on both datasets.

\end{abstract}

\section{Introduction}
\label{intro}

Scene classification is a computer vision task that predicts the scene category by analyzing the background and target objects \cite{Yi_SceneSeg_2019IEEEAcess}. It remains as an open and challenging research topic widely applied \cite{survey_2020} in video surveillance, human-computer interaction.
In recent years, indoor scene classification approaches have also been integrated into robotics perception modules \cite{robotics_application, Pereira_roman22} to recognize the surrounding environment.

The same scene category can have multiple furniture configurations, points of view, and scene area, so obtaining a feature space that covers all the aforementioned conditions may be difficult \cite{gsf2appV2}.   
This intra-category variation issue needs to be taken into account, especially when the same category changes so much that only a few common patterns can be extracted. Moreover, as the number of scene categories increases, more appearance similarities are shared between scene categories, which leads to a negative impact on the inter-scene boundaries, making it difficult to obtain a correct scene category prediction \cite{lopezcifuentes_SceneSeg_2020PR}. This issue is designated by inter-category ambiguity.

\begin{figure*}[tb]
	\centering
	\includegraphics[width=\linewidth]{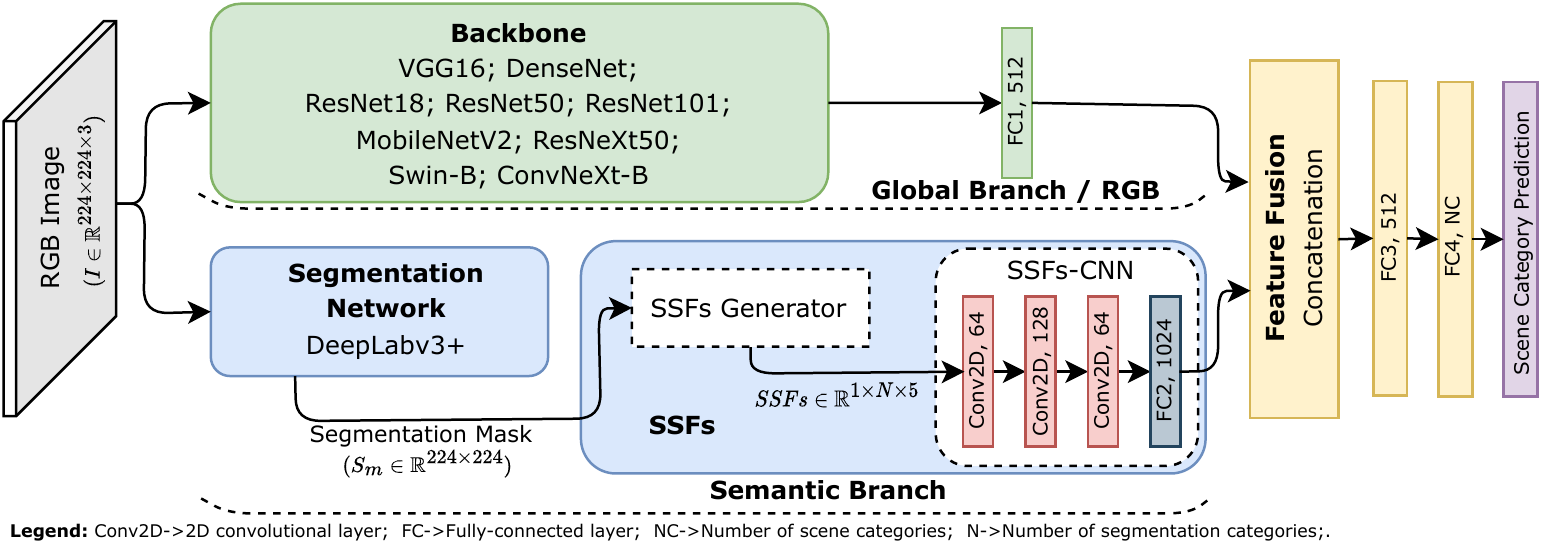}
	\caption{Overview of the proposed GS$^2$F$^2$App. The global branch (green) uses a state-of-the-art CNN to extract global features. The semantic branch (blue) generates a semantic segmentation mask, from which SSFs are extracted, which are exploited by the SSF-CNN. Both branches converge to a feature fusion module (yellow), leading to a scene class prediction.}
	\label{fig:gs2f2app_v2_overview}	
\end{figure*}

To identify an indoor scene, people focus on the objects available in it, and also on how the objects are distributed across the scene \cite{df2net_2018}. Therefore, inspired by how people identify an indoor scene, recent works \cite{gsf2appV2,lopezcifuentes_SceneSeg_2020PR, song_oor_2020, mapnet_2019} have been trying to overcome the intra-category variation and inter-category ambiguity issues by exploiting semantic information, such as object-related information and object correlations across the scene. To obtain semantic information, such works have relied on object detector techniques, that predict object categories and objects' location as 2D bounding boxes, to locate the objects in the scene. 

Conversely, semantic information can be gathered by using semantic segmentation approaches \cite{lopezcifuentes_SceneSeg_2020PR} that provide semantic segmentation masks containing a pixel-level classification of an image, i.e., at each pixel is assigned a segmentation-category (e.g. object, wall, floor). Hence, through semantic segmentation masks, a well-defined segmentation-category's shape is obtained. Semantic segmentation masks can lead to a more accurate object-based spatial distribution over the indoor scene.

This paper presents a novel approach, segmentation-based semantic features (SSFs), that, considering a semantic segmentation mask, allows to obtain a semantic feature representation that depicts how segmentation-categories are spread across the scene. Such features represent, per segmentation-category, the pixel count, the 2D average position, and their respective standard deviation values. Thus, the proposed SSFs provide a 2D spatial layout of the segmentation-categories across the scene, leading to a more meaningful semantic feature representation. 
Moreover, a two-branch CNN-based Global and Segmentation-based Semantic Feature Fusion Approach (GS$^2$F$^2$App), that exploits CNN-based global features and the proposed SSFs extracted from semantic segmentation masks, is also proposed (see Fig.\,\ref{fig:gs2f2app_v2_overview}). Both branches' output features are concatenated for scene prediction. Semantic segmentation masks are generated by the DeepLabv3+ semantic segmentation network \cite{chen2018encoder}. The proposed GS$^2$F$^2$App was evaluated on two indoor scene benchmark datasets: SUN RGB-D \cite{sun_dataset} and NYU Depth V2 \cite{nyu_dataset}. The main contributions of this work can be summarized as follows:

\begin{itemize}
    \item A novel segmentation-based approach, which extracts SSFs representing how segmentation-categories are distributed in the scene;
    
    \item GS$^2$F$^2$App: a two-branch network that exploits global features extracted from RGB images and the SSFs;

    \item A comprehensive study to evaluate the influence of each feature that makes up the proposed SSFs (pixel count, the 2D average position, and their respective standard deviation values) may have in representing indoor scenes, is provided. 
    
    \item The GS$^2$F$^2$App was evaluated on the SUN RGB-D dataset \cite{sun_dataset} and on the NYU Depth V2 dataset \cite{nyu_dataset}. To the best of our knowledge, state-of-the-art results were achieved on both datasets.
\end{itemize}

\section{Related Work}
\label{sec:related_work}

The rise of deep learning techniques has piqued researchers' interest in indoor scene classification. As a result, several methodologies and approaches have been developed to tackle this challenging problem \cite{survey_2020}. In the literature, irrespective of the techniques used in each work, the majority of them adhere to a three-stage pipeline \cite{survey_2020}: feature extraction, transformation and aggregation, and scene category prediction. Nevertheless, due to the variation and ambiguity found in indoor scenes, extracted CNN-based global features are insufficient to represent them \cite{xiong_2019}. To overcome such issues, new feature representations mainly focused on multi-modal and semantic features have been researched.

Since depth data allow the extraction of some semantic cues regarding objects' shape, multi-modality approaches with a greater focus on RGB-D modalities have been proposed \cite{df2net_2018, mapnet_2019, caglayan_2022}. Li \textit{et al.} \cite{df2net_2018} proposed a feature correlation approach that learns how to correlate features extracted from each modality. Later, such work was improved in \cite{mapnet_2019} by integrating RGB-D semantic cues obtained by a region proposal technique. Caglayan \textit{et al.} \cite{caglayan_2022} mapped CNN-based RGB-D features extracted at multiple levels into high-level representations through a fully randomized structure of recursive neural networks. Results attained by multi-modality approaches showed that a more discriminative feature representation of indoor scenes is obtained compared with the one achieved by using a single modality. However, such approaches only relied on shape cues and did not consider object categories and their distribution in the scene, which can be important to disentangle scene predictions.

Aiming to get more meaningful information about the scene that can lead to better scene representations, in addition to CNN-based global features, object-based features have been proposed in \cite{gsf2appV2, song_oor_2020, song_2017, cheng_sdo_2018, Zhou_iros_2021}.
Pereira \textit{et al.} \cite{gsf2app} exploited CNN-based global features with the occurrence of objects recognized in the scene. Later, an improvement by exploiting inter-object distance relationships was proposed in \cite{gsf2appV2}. Song \textit{et al.} \cite{song_2017} proposed scene spatial layout representations generated using an object-to-object correlation method. An improvement presenting two object-based image representations that exploit the relationships between co-occurring objects and relationships from a sequential occurrence of objects was proposed in \cite{song_oor_2020}. Despite the promising results achieved by such approaches, object-based features were generated from object detector techniques, which fail to provide accurate information about the whole scene (e.g., floor, walls) or pixel-level object's shape that can play an important role in disentangling the scene predictions.

In addition to object detector techniques, alternative approaches for extracting object-based features involve utilizing semantic segmentation masks \cite{lopezcifuentes_SceneSeg_2020PR, Perdiguero_SceneSeg_2018Iros, Ahmed_SceneSeg_2020Sensors}. Ahmed \textit{et al.} \cite{Ahmed_SceneSeg_2020Sensors} proposed multi-object categorizations to obtain object representations within the scenes. Herranz-Perdiguero \textit{et al.} \cite{Perdiguero_SceneSeg_2018Iros} used the segmentation mask to locate recognized objects in the scene and generate pixel distributions of objects using a histogram-like approach. Furthermore, López-Cifuentes \textit{et al.} \cite{lopezcifuentes_SceneSeg_2020PR} introduced a two-branch network that simultaneously exploits global features from an RGB image and local semantic features from a semantic segmentation mask. 
Driven by the positive aspects of object-based features extracted from semantic segmentation masks and considering that the aforementioned object-based and segmentation-based techniques do not exploit the 2D spatial layout of objects within indoor scenes, this paper introduces the SSFs that encode the 2D spatial layout of segmentation-categories across the scene.
The herein proposed approach is an improvement over a previous method \cite{gsf2appV2}, which introduced inter-object distance relationships from 2D bounding boxes acquired through an object detector technique. The GS$^2$F$^2$App shares similarities with \cite{lopezcifuentes_SceneSeg_2020PR}, as both utilize CNN-based global features and semantic features through a semantic segmentation mask. 
However, while the semantic features in \cite{lopezcifuentes_SceneSeg_2020PR} encode scene object probabilities, the GS$^2$F$^2$App captures the distribution of segmentation-categories across the scene.


\section{Methodology}
\label{sec:methodology}

Figure \ref{fig:gs2f2app_v2_overview} provides an overview of the proposed GS$^2$F$^2$App, consisting of two branches: the global branch and the semantic branch. The output features from both branches undergo a feature fusion stage, wherein additional feature correlations are learned, leading to the final scene category prediction.

\subsection{Semantic Branch}

A scene category is depicted by the available semantic information, which comprises the objects scattered throughout the scene and their relationships. Therefore, aiming to obtain meaningful semantic information about indoor scenes, SSFs extracted from semantic segmentation masks are proposed. To predict semantic segmentation masks from RGB images, the DeepLabv3+ \cite{chen2018encoder} encoder-decoder-based semantic segmentation network is used.

\subsection*{Segmentation-based Semantic Features}

The proposed SSFs encode a 2D spatial layout of the segmentation-categories over an indoor scene, representing how each segmentation-category is scattered across the scene. Thus, SSFs are composed of the following features:
i) the pixel count ($P_C$) of each segmentation-category; ii) the 2D average position ($I_{\mu_x}$, $I_{\mu_y}$) per segmentation-category; iii) 2D standard deviation values ($I_{\sigma_x}$, $I_{\sigma_y}$) based on each 2D pixel coordinates corresponding to each segmentation-category.

Let $L$ be the number of segmentation-categories, $n\in [1,L]$ an index identifying an arbitrary segmentation-category, and $S_m$ a semantic segmentation mask with size $w\times h$, where the pixel value $S_m(i,j) \in [1,L]$ represents an index corresponding to a segmentation-category at the indexes $i = [1,...,h]$, and $j=[1,..,w]$.

\textbf{Pixel count:} represents how much a specific segmentation-category is present in the scene, obtained as follows:
\begin{equation}
    \label{eq:pixel_number}
    P_C(n) = \sum_{i=1}^{h}{\sum_{j=1}^{w}{}} f_C(i,j,n),
\end{equation}
with
\begin{equation}
    f_C(i,j,n) = \left\{\begin{matrix} 
		1 & \textup{if }S_m(i,j) = n \\ 
		0 & \textup{otherwise }
	\end{matrix}\right.
 .
\end{equation}
A normalized value of $P_C(n)$ is obtained by dividing it by the total number of pixels, $P_C'(n)$, expressed as follows:
\begin{equation}
    \label{eq:norm_pixel_number}
    P_C'(n) = \frac{P_C(n)}{h\times w}. 
\end{equation}

\textbf{2D average position:} provides spatial information regarding each segmentation-category, representing the average location (in the image space) where each segmentation-category is mainly present in the scene.
It is calculated by using the segmentation mask's indexes ($i,j$) that correspond to a specific segmentation-category ($n$), as follows:
\begin{equation}
    \label{eq:I_mu_x}
    I_{\mu_x}(n) = \frac{\sum_{i=1}^{h}{\sum_{j=1}^{w}{}} j \times f_C(i,j,n) }{P_C(n)},
\end{equation}
 \begin{equation}
    \label{eq:I_mu_y}
    I_{\mu_y}(n) = \frac{\sum_{i=1}^{h}{\sum_{j=1}^{w}{}} i \times f_C(i,j,n) }{P_C(n)}.
\end{equation}
The 2D average position allows to obtain a meaningful 2D spatial layout of the segmentation-categories across the scene. The 2D average position values are normalized as follows: 

\begin{equation}
    \label{eq:norm_I_mu_x}
    I_{\mu_x}'(n) = \frac{1}{w} I_{\mu_x} (n),
\end{equation}
\begin{equation}
    \label{eq:norm_I_mu_y}
    I_{\mu_y}'(n) = \frac{1}{h} I_{\mu_y} (n).
\end{equation}

\begin{figure}[tb]
    \centering 
    \includegraphics[trim={0cm 0.14cm 0cm 0.0cm}, clip, width=\linewidth]{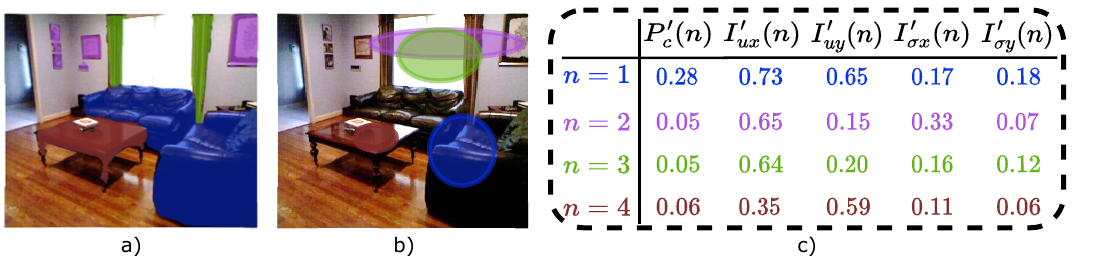}
    \caption{Visual representation of the proposed SSFs, being: a) RGB image with overlaid semantic segmentation mask; b) RGB image overlaid with a 2D spatial representation of four segmentation-categories (''sofa''(n=1), ''painting''(n=2), ''curtain''(n=3), ''table''(n=4)) through the segmentation mask. The 2D spatial representation was obtained by applying the 2D standard deviation values of each 2D position regarding each segmentation-category centered in the 2D average position per segmentation-category; c) Obtained SSFs' values.
    The RGB and segmentation mask images were taken from the NYU Depth Dataset V2 \cite{nyu_dataset}.}
    \label{fig:StatisticsFeatures}	
\end{figure}

\textbf{2D standard deviation values:} provide spatial information regarding each segmentation-category, representing how close or apart pixels of the same segmentation-category are. The 2D standard deviation values are calculated using the segmentation mask's indexes ($i,j$) that correspond to a specific segmentation-category ($n$), as follows:
\begin{equation}
    \label{eq:I_sigma_x}
    I_{\sigma_x}(n) = \sqrt{\frac{\sum_{i=1}^{h}{\sum_{j=1}^{w}{}} (j-I_{\mu_x}(n))^2 \times f_C(i,j,n)}{P_C(n)}}, 
\end{equation}
\begin{equation}
    \label{eq:I_sigma_y}
    I_{\sigma_y}(n) = \sqrt{\frac{\sum_{i=1}^{h}{\sum_{j=1}^{w}{}} (i-I_{\mu_y}(n))^2 \times f_C(i,j,n)}{P_C(n)}}. 
\end{equation}
A high standard deviation value means that the segmentation-category is spread out across the scene, while a low standard deviation value means that the segmentation-category is concentrated in a small region of the scene, as shown by the ''painting'' and ''table'' segmentation-categories in Fig.\,\ref{fig:StatisticsFeatures}, respectively. The 2D standard deviation values, along with the 2D average position provide relevant spatial information of the segmentation-categories across the scene, allowing to obtain a meaningful 2D spatial layout of the segmentation-categories over the scene. 
Obtained 2D standard deviation values are normalized as follows: 

\begin{equation}
    \label{eq:norm_I_sigma_x}
    I_{\sigma_x}'(n) = \frac{1}{w} I_{\sigma_x}(n),
\end{equation}
\begin{equation}
    \label{eq:norm_I_sigma_y}
    I_{\sigma_y}'(n) = \frac{1}{h} I_{\sigma_y}(n).
\end{equation}

All the aforementioned normalized features are organized under a matrix form, SSFs, represented as follows:
\begin{equation}
	\label{isc:matrix:ssf}
	SSFs = 
	\begin{bmatrix}
		P_C'(1) & I_{\mu_x}'(1) & I_{\mu_y}'(1) & I_{\sigma_x}'(1) &             I_{\sigma_y}'(1) \\
		P_C'(2) & I_{\mu_x}'(2) & I_{\mu_y}'(2) & I_{\sigma_x}'(2) & I_{\sigma_y}'(2) \\
		\vdots & \vdots & \vdots & \vdots & \vdots \\
		P_C'(L) & I_{\mu_x}'(L) & I_{\mu_y}'(L) & I_{\sigma_x}'(L) & I_{\sigma_y}'(L) \\
	\end{bmatrix}  
	\in \mathbb{R}^{L\times 5}.
\end{equation}

\subsection*{SSFs-CNN Architecture}

To further exploit the correlation and non-linear patterns among the proposed SSFs, a CNN architecture, SSFs-CNN, as shown in Fig. \ref{fig:gs2f2app_v2_overview}, is proposed. It is composed of three 2D convolutional layers (with 64, 128, and 64 output channels, respectively) followed by one fully-connected (FC) layer (with 1024 output features). Convolutional layers use a kernel size of 3, a stride value of 1, and a padding value of 1. All convolutional and FC layers use the ReLU activation function.

\subsection{Global Branch}

The proposed SSFs focus solely on the spatial characterization of segmentation-categories. While semantic-based features are essential to represent an indoor scene, global aspects of the scene such as the background or building
structure/characteristics, can also play a significant role in scene representation. Hence, global features are extracted and learned from RGB images through a state-of-the-art deep-learning-based feature extraction approach (backbone), as depicted in the top branch of the proposed GS$^2$F$^2$App (see Fig. \ref{fig:gs2f2app_v2_overview}). Furthermore, the global branch includes an FC layer that receives the backbone's output feature space as input. The FC layer utilizes the ReLU activation function.
To exploit the effectiveness of the different deep learning-based feature extraction techniques in indoor scene classification tasks, and also to show that the proposed SSFs do not have a global feature extraction dependency, the feature extraction layers of the following six state-of-the-art CNNs, also designated as backbone networks, were assessed: VGG16 \cite{vgg}, ResNet18-50-101 \cite{resnet}, DenseNet \cite{densenet}, MobileNetV2 \cite{mobileNetv2}, ResNeXt50 \cite{resNext_2017}, and ConvNeXt-B \cite{convNext_2022}. Moreover, the transformer-based Swin-B \cite{swin_2021} network is also considered as a backbone network.

\subsection{Feature Fusion}

To exploit the correlation between global and semantic features, the GS$^2$F$^2$App uses a two-step learning technique similar to the one employed in \cite{gsf2appV2}.
In the first learning step, an RGB-only base model is obtained, i.e., only the backbone network's weights, as well as the FC1's weights, are trained. In the second learning step, the weights trained in the previous learning step are frozen, and the remaining weights (SSFs-CNN, FC3, and FC4) are trained. 
To merge features from global and semantic branches, GS$^2$F$^2$App employs a concatenation feature fusion approach. It consists in obtaining a feature fusion vector ($F_F$) by concatenating the output's feature vectors of global and semantic branches, expressed as follows:
\begin{equation}
    \label{eq:feature_concat}
    \begin{split}
        F_F & = F_{Global} \ \mathbin\Vert F_{SSFs} = [F_{Global} \ \  F_{SSFs}], \\
            & = [u_1,\ \hdots ,\ u_{l_G}, \  v_1,\ \hdots ,\ v_{l_{SSFs}} ],
    \end{split}
\end{equation}
where $\mathbin\Vert$ represent the concatenation operator, $F_{Global} = [u_1,\ \hdots \ ,u_{l_G}],\  u_{i=1:l_G} \in \mathbb{R}$ the global branch's output feature vector, $F_{SSFs} = [v_1,\ \hdots \ ,v_{l_{SSFs}}],\ v_{i=1:l_{SSFs}} \in \mathbb{R}$ the SSF-CNN's output feature vector, being $l_G$ and $l_{SSFs}$ the total number of global and SSFs features, respectively.

\section{Experiments}
\label{sec:experiments}

The proposed approach was evaluated on two scene classification datasets: SUN RGB-D \cite{sun_dataset}, and NYU Depth v2 \cite{nyu_dataset}.

\subsection{Datasets}

\textbf{\textit{1) SUN RGB-D Dataset:}} It has available 10355 RGB-D image pairs captured from different cameras: Kinect v1, Kinect v2, RealSense, and Asus Xtion. Following the public split in \cite{sun_dataset}, there are 4845 training images and 4659 testing images. The dataset contains 19 indoor scene categories.

\textbf{\textit{2) NYU Depth v2 (NYUv2):}} It contains 1449 RGB-D images distributed into 27 scene categories. However, following the benchmark split \cite{nyu_data_split}, the original scene categories were reorganized into 10 scene categories (9 most common and "other"). The dataset split has 795 training and 654 testing images.

\subsection{Implementation Details}

The DeepLabv3+ \cite{chen2018encoder} encoder network's weights were initialized with an ImageNet pre-trained model and fine-tuned over 100 epochs using the AdamW optimizer with a learning rate of 0.001 and a weight decay of 0.05. The scope of this work does not cover the assessment of the generalization capabilities of the segmentation network. Instead, the main focus is on assessing the performance of the proposed semantic feature representation in the GS$^2$F$^2$App. As presented in \cite{Seg_benchmark}, the state-of-the-art segmentation result achieved on the SUN RGB-D and NYU depth v2 datasets is approximately 50\%. Therefore, to obtain segmentation data with higher performance for evaluating the proposed approach, both the training and testing sets of the datasets were used during the training phase.
In the first learning step of the GS$^2$F$^2$App, the weights of the backbone network were initialized using an ImageNet pre-trained model and fine-tuned over 100 epochs. The second learning step combines the CNN-based global features with the SSFs over 100 epochs. In both learning steps, the ADAM optimizer with a learning rate of $10^{-4}$, a weight decay rate of $0.0005$, and a mini-batch size of 32 were used.
All experiments were implemented using the Python 3.10.4 programming language and the PyTorch framework (version 1.11.0), performed using an Nvidia RTX 3090 GPU, 64GB RAM, and an AMD Ryzen 9-5900X-@-3.7 GHz. All RGB images and segmentation masks were resized to $224\times 224$.

\subsection{Results: Comparison with SOTA Methods}

To demonstrate the efficacy of the proposed approach, we compare the achieved results with those obtained from state-of-the-art methods, as shown in Table \ref{tab:state_of_the_art_results}. The proposed GS$^2$F$^2$App achieved 62.3\% and 77.8\% accuracy values on the SUN RGB-D and NYU Depth datasets, respectively, which is, to the best of our knowledge, the highest reported state-of-the-art results. Promising overall performances, using the RGB-only or RGB-D modalities, were also reached by \cite{gsf2appV2,song_oor_2020,mapnet_2019,caglayan_2022,Montoro_2021,Xiong_2021,Ayub_centroid_2020,Seichter_2022}. 
In \cite{gsf2appV2}, inter-object distance relationships, encapsulating the proximity or separation of object categories within the scene, were introduced. Ayub \& Wagner \cite{Ayub_centroid_2020} proposed a centroid-based concept learning that generates clusters and centroid pairs for different scene classes.
Close performances were also reached by \cite{song_oor_2020,mapnet_2019,Xiong_2021} through exploiting global and local semantic features available in the scenes. Caglayan \textit{et al.} \cite{caglayan_2022} used a CNN to extract visual features at multiple levels, which were subsequently mapped into high-level representations using a recursive neural network. Montoro \& Hidalgo \cite{Montoro_2021} exploited 2D visual texture features and 3D object-related geometric features. 
Seichter \textit{et al.} \cite{Seichter_2022} introduced an RGB-D multi-task network for scene classification, semantic and instance segmentation, and instance orientation. Such an approach enables the backbone network to learn how to extract descriptive features from RGB-D image pairs, including global and semantic features, resulting in a promising indoor scene classification performance.
Conversely the GS$^2$F$^2$App captures a spatial distribution of segmentation-categories in a 2D format from semantic segmentation masks. The achieved results demonstrate that when this distribution is combined with CNN-based global features, a superior feature representation of indoor scenes is achieved.

\setlength{\tabcolsep}{7.3pt}
\begin{table}[!tb]
    \centering
    \footnotesize
    \caption{Accuracy(\%) comparison with state-of-the-art methods.}
    \label{tab:state_of_the_art_results}
    \begin{tabular}{ccccc}
    \noalign{\hrule height 1pt} \hline
    \multirow{3}{*}{Method} & \multicolumn{4}{c}{Dataset}                                           \\ \cline{2-5} 
                            & \multicolumn{2}{c}{SUN RGB-D}             & \multicolumn{2}{c}{NYUv2} \\ \cline{2-5} 
                            & RGB  & RGB-D                              & RGB         & RGB-D       \\ \hline
    Gupta \textit{et al.} \cite{nyu_data_split}     & -    & \multicolumn{1}{c|}{-}             & 58.0        & -           \\
    Song \textit{et al.} \cite{song_2017}      & -    & \multicolumn{1}{c|}{54.0}          & 57.3        & 66.9        \\
    Li \textit{et al.} \cite{df2net_2018}         & 46.3 & \multicolumn{1}{c|}{54.6}          & 61.1        & 65.4        \\
    Du \textit{et al.} \cite{du_2018}        & 42.6 & \multicolumn{1}{c|}{53.3}          & 53.7        & 67.5        \\
    Xiong \textit{et al.} \cite{xiong_2019}      & -    & \multicolumn{1}{c|}{55.9}          & 53.5        & 67.8        \\
    Li \textit{et al.} \cite{mapnet_2019}        & -    & \multicolumn{1}{c|}{56.2}          & -           & 67.7        \\
    Song \textit{et al.} \cite{song_oor_2020}         & 50.5 & \multicolumn{1}{c|}{55.5}          & 64.2        & 67.4        \\
    Ayub \& Wagner \textit{et al. \cite{Ayub_centroid_2020}} & 48.8 & \multicolumn{1}{c|}{59.5} & 66.4          & 70.9     \\
    Xiong \textit{et al.} \cite{Xiong_2021}         & 55.3 & \multicolumn{1}{c|}{57.3}             & 61.2        & 69.3           \\ 
    Pereira \textit{et al.} \cite{gsf2appV2}       & 58.2      & \multicolumn{1}{c|}{-}             & 73.1            & -       \\
    Montoro \& Hidalgo \cite{Montoro_2021}       & 56.4      & \multicolumn{1}{c|}{58.6}             & 67.8      & 75.1       \\
    Caglayan \textit{et al.} \cite{caglayan_2022}     & 58.5 & \multicolumn{1}{c|}{60.7} & -           & -           \\
    Seichter \textit{et al.} \cite{Seichter_2022}   & - & \multicolumn{1}{c|}{\textbf{61.8}} & -           & \textbf{76.5}     \\ \hline
    GS$^2$F$^2$App (ours)                    & \textbf{62.3}      & \multicolumn{1}{c|}{-}             & \textbf{77.8}            & -       \\ \hline
     \noalign{\hrule height 1pt} \hline
    \end{tabular}
\end{table}

\subsection{Results: Ablation Study}

To evaluate the influence that the segmentation model performance may have in the proposed approach, since the quality of the extracted SSFs is correlated with the segmentation model performance, three different segmentation models (S20, S50, and S75), considering different segmentation mean Intersection over Union (mIoU) values, were used.
S20/S50/S75 correspond to segmentation models with mIoU values of approximately 20\%/50\%/75\%, respectively. Taking into account the state-of-the-art in segmentation results \cite{Seg_benchmark}, the S50 model was selected as the benchmark.

\begin{figure}[!tb]
    \centering
    \includegraphics[width=\linewidth]{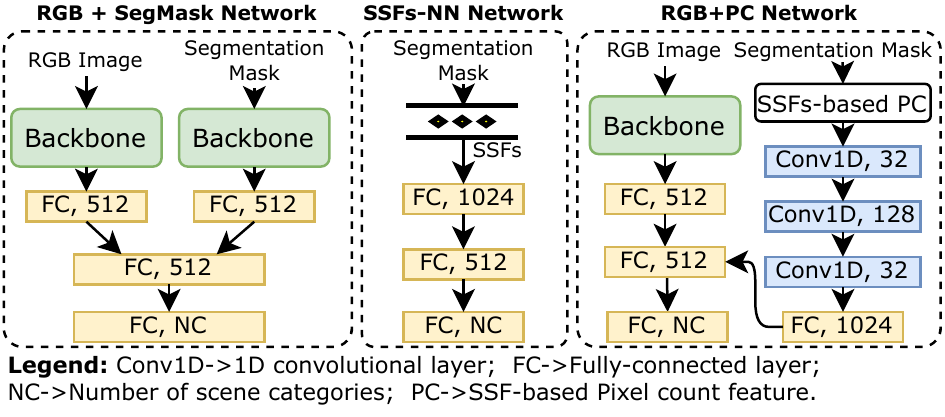}
    \caption{Network architectures used in the ablation study. The RGB+SegMask network independently extracts DL features from an RGB image and a semantic segmentation mask that are concatenated for scene prediction. The SSFs-NN network uses fully-connected layers to exploit correlations from the proposed SSFs. The RGB+PC network extracts CNN-based features from an RGB image and uses 1D convolutional layers to exploit the SSFs-based PC features. Both extracted features are concatenated for scene prediction.}
	\label{fig:ablation}	
\end{figure}

To further assess the effectiveness of the proposed GS$^2$F$^2$App, the following intermediate evaluations were also performed: global features-only (RGB); global features extracted from the semantic segmentation mask (SegMask); 
semantic features-only exploited by convolutional layers convolutional layers (SSFs-CNN) or fully-connected layers (SSFs-NN), SSFs-based pixel count features (PC-CNN, PC-NN), SSFs-based 2D average position features (AP-CNN, AP-NN), SSFs-based 2D standard deviation features (SD-CNN, SD-NN), and combinations between SSFs-based features (e.g., PC\&AP-CNN, and AP\&SD-NN); global features combined with SSFs-based features exploited by convolutional layers (e.g., RGB+PC and RGB+PC\&SD); global features combined with segmentation mask-based global features (RGB+SegMask); the proposed approach using fully-connected (FC) layers to exploit the SSFs (GS$^2$F$^2$App-NN), and global features extracted by different backbone networks.
To accomplish such intermediate evaluations, three network architectures, as shown in Fig. \ref{fig:ablation}, were developed. 
The RGB+SegMask network combines deep-learning-based (DL) features extracted from an RGB image and a segmentation mask. The SSFs-NN network exploits the SSFs through fully-connected layers.
Conversely, in the evaluation of the SSFs-based PC-only features, 1D convolutional layers were employed, given that the PC features are encoded as a vector, as illustrated in the RGB+PC network (refer to Fig. \ref{fig:ablation}).
For the remaining evaluations, the GS$^2$F$^2$App architecture was used.

\setlength{\tabcolsep}{7.6pt}
\begin{table}[!tb]
    \centering
    \footnotesize
    \caption{Accuracy (\%) values for the semantic features-only evaluation.}
    \label{tab:ssf_results}
    \begin{tabular}{ccccccc}
    \noalign{\hrule height 1pt} \hline 
    \multirow{3}{*}{Feature-Network} & \multicolumn{6}{c}{Dataset}                                      \\ \cline{2-7} 
                                     & \multicolumn{3}{c|}{SUN RGB-D}       & \multicolumn{3}{c}{NYUv2} \\ \cline{2-7} 
                                     & S20 & S50 & \multicolumn{1}{c|}{S75} & S20     & S50    & S75    \\ \hline
        PC-CNN      & 33.1 & 42.0 & \multicolumn{1}{c|}{43.2} & 54.4 & 66.0 & 68.4 \\
        PC-NN       & 32.7 & 41.8 & \multicolumn{1}{c|}{42.5} & 54.9 & 66.2 & 68.0 \\
        AP-CNN      & 28.7 & 36.1 & \multicolumn{1}{c|}{37.5} & 53.3 & 64.4 & 66.5 \\
        AP-NN       & 29.4 & 33.9 & \multicolumn{1}{c|}{34.5} & 53.3 & 65.0 & 66.9 \\
        SD-CNN      & 31.1 & 37.8 & \multicolumn{1}{c|}{38.5} & 54.4 & 66.3 & 68.4 \\
        SD-NN       & 31.8 & 36.9 & \multicolumn{1}{c|}{37.8} & 54.2 & 66.6 & 68.1 \\
        AP\&SD-CNN  & 31.8 & 39.2 & \multicolumn{1}{c|}{40.9} & 54.7 & 66.6 & 68.6 \\
        AP\&SD-NN   & 31.4 & 37.8 & \multicolumn{1}{c|}{38.3} & 55.0 & 66.5 & 68.4 \\
        PC\&AP-CNN  & 33.4 & 42.7 & \multicolumn{1}{c|}{43.6} & 54.6 & 66.4 & 68.7 \\
        PC\&AP-NN   & 33.5 & 42.5 & \multicolumn{1}{c|}{43.0} & 55.0 & 66.6 & 69.0 \\
        PC\&SD-CNN  & 34.1 & 43.2 & \multicolumn{1}{c|}{44.1} & 55.3 & 67.0 & 69.1 \\
        PC\&SD-NN   & 34.3 & 42.7 & \multicolumn{1}{c|}{44.1} & 55.0 & 66.9 & 69.3 \\
        SSFs-CNN    & 34.6 & \textbf{43.5} & \multicolumn{1}{c|}{\textbf{44.5}} & 56.4 & 67.3 & 70.0 \\ 
        SSFs-NN     & \textbf{34.9} & 43.3 & \multicolumn{1}{c|}{43.3} & \textbf{56.8} & \textbf{67.5} & \textbf{70.1} \\ \hline
        \noalign{\hrule height 1pt} \hline
    \end{tabular}
\end{table}

\setlength{\tabcolsep}{5pt}
\begin{table*}[!tb]
    \centering
    \footnotesize
    \caption{Ablation study on the SUN RGB-D dataset.}
    \label{tab:sun_results}
    \begin{tabular}{lccccccccccccccc}
    \noalign{\hrule height 1pt} \hline
                & \multicolumn{15}{c}{Accuracy (\%)}                                                        \\ \cline{2-16} 
    \multirow{2}{*}{Backbone} &
        \multirow{2}{*}{RGB} &
      \multicolumn{2}{c}{SegMask} &
      \multicolumn{2}{c}{\begin{tabular}[c]{@{}c@{}}RGB +\\ SegMask\end{tabular}} &
      \begin{tabular}[c]{@{}c@{}}RGB+\\ PC\end{tabular} &
      \begin{tabular}[c]{@{}c@{}}RGB+\\ AP\end{tabular} &
      \begin{tabular}[c]{@{}c@{}}RGB+\\ SD\end{tabular} &
      \begin{tabular}[c]{@{}c@{}}RGB+\\ AP\&SD\end{tabular} &
      \begin{tabular}[c]{@{}c@{}}RGB+\\ PC\&AP\end{tabular} &
      \begin{tabular}[c]{@{}c@{}}RGB+\\ PC\&SD\end{tabular} &
      \multicolumn{2}{c}{GS$^2$F$^2$App} &
      \multicolumn{2}{c}{GS$^2$F$^2$App-NN} \\ \cline{3-16} 
                & & S50 & S75 & S50 & S75 & S50 & S50 & S50 & S50 & S50 & S50 & S50 & S75 & S50 & S75 \\ \hline \hline
    VGG16 \cite{vgg}      & 52.4 & 44.2 & 45.5 & 56.3 & 56.6 & 55.9 & 56.3 & 56.0 & 56.6 & 56.7 & 56.7 & 56.9 & 56.9 & 56.5 & 56.8 \\
    ResNet18 \cite{resnet}   & 56.0 & 49.1 & 51.5 & 58.2 & 58.6 & 57.8 & 58.1 & 57.8 & 58.5 & 58.3 & 58.4 & 58.7 & 58.7 & 58.7 & 58.5 \\
    ResNet50 \cite{resnet}   & 56.8 & 49.3 & 51.5 & 59.2 & 59.5 & 59.0 & 59.2 & 59.0 & 59.6 & 59.5 & 59.5 & 59.8 & 59.6 & 59.5 & 59.4 \\
    ResNet101 \cite{resnet}  & 56.3 & 49.1 & 50.7 & 58.6 & 59.0 & 58.4 & 58.6 & 58.4 & 58.9 & 59.0 & 59.0 & 59.2 & 59.3 & 59.1 & 59.2 \\
    DenseNet \cite{densenet}   & 57.1 & 50.7 & 52.8 & 58.4 & 58.9 & 58.0 & 58.2 & 58.1 & 58.5 & 58.5 & 58.7 & 58.9 & 59.1 & 58.8 & 58.9 \\
    MobileNetV2 \cite{mobileNetv2} & 56.7 & 49.8 & 51.8 & 56.1 & 56.4 & 55.3 & 55.6 & 55.4 & 56.3 & 56.1 & 55.9 & 56.5 & 55.9 & 56.5 & 56.3 \\ 
    ResNeXt50 \cite{resNext_2017}  & 56.9 & \textbf{51.2} & 53.3 & 61.5 & 61.8 & 61.4 & 61.5 & \textbf{61.4} & 61.8 & 61.7 & 61.6 & 62.1 & 62.1 & \textbf{62.0} & 61.8 \\ 
    Swin-B \cite{swin_2021}     & 57.3 & 50.6 & 52.5 & 60.6 & 60.9 & 60.5 & 60.5 & 60.3 & 60.9 & 60.7 & 60.7 & 61.1 & 61.6 & 60.8 & 60.8 \\ 
    ConvNeXt-B \cite{convNext_2022} & \textbf{58.3} & 51.0 & \textbf{53.8} & \textbf{61.9} & \textbf{62.1} & \textbf{61.5} & \textbf{61.6} & 61.3 & \textbf{62.1} & \textbf{61.8} & \textbf{61.7} & \textbf{62.3} & \textbf{62.4} & 61.9 & \textbf{62.1} \\ 
    \noalign{\hrule height 1pt} \hline
    \end{tabular}
\end{table*}

\setlength{\tabcolsep}{5pt}
\begin{table*}[!tb]
    \centering
    \footnotesize
    \caption{Ablation study on the NYU depth dataset v2.}
    \label{tab:nyu_results}
    \begin{tabular}{lccccccccccccccc}
    \noalign{\hrule height 1pt} \hline
                & \multicolumn{15}{c}{Accuracy (\%)}                                                        \\ \cline{2-16} 
    \multirow{2}{*}{Backbone} &
      \multirow{2}{*}{RGB} &
      \multicolumn{2}{c}{SegMask} &
      \multicolumn{2}{c}{\begin{tabular}[c]{@{}c@{}}RGB +\\ SegMask\end{tabular}} &
      \begin{tabular}[c]{@{}c@{}}RGB+\\ PC\end{tabular} &
      \begin{tabular}[c]{@{}c@{}}RGB+\\ AP\end{tabular} &
      \begin{tabular}[c]{@{}c@{}}RGB+\\ SD\end{tabular} &
      \begin{tabular}[c]{@{}c@{}}RGB+\\ AP\&SD\end{tabular} &
      \begin{tabular}[c]{@{}c@{}}RGB+\\ PC\&AP\end{tabular} &
      \begin{tabular}[c]{@{}c@{}}RGB+\\ PC\&SD\end{tabular} &
      \multicolumn{2}{c}{GS$^2$F$^2$App} &
      \multicolumn{2}{c}{GS$^2$F$^2$App-NN} \\ \cline{3-16} 
                &  & S50 & S75 & S50 & S75 & S50 & S50 & S50 & S50 & S50 & S50 & S50 & S75 & S50 & S75 \\ \hline \hline
    VGG16 \cite{vgg}      & 63.2 & 60.3 & 61.8 & 67.1 & 67.4 & 65.2 & 68.3 & 65.2 & 71.2 & 70.4 & 71.5 & 72.0 & 71.8 & 66.6 & 66.3 \\
    ResNet18 \cite{resnet}   & 68.0 & 66.5 & 67.5 & 71.5 & 72.7 & 69.4 & 73.2 & 69.4 & 74.3 & 74.0 & 73.7 & 75.0 & 75.3 & 73.1 & 73.2 \\
    ResNet50  \cite{resnet}  & 70.8 & 69.3 & 70.2 & 74.5 & 74.7 & 72.9 & 73.8 & 72.9 & 74.8 & 74.6 & 74.3 & 75.5 & 75.8 & 73.7 & 74.0 \\
    ResNet101 \cite{resnet}  & 70.7 & 70.1 & 71.1 & 74.4 & 74.9 & 72.1 & 72.8 & 72.2 & 74.0 & 73.9 & 73.4 & 74.9 & 75.8 & 73.5 & 73.5 \\
    DenseNet \cite{densenet}   & 70.8 & 68.6 & 70.0 & 73.2 & 74.3 & 71.5 & 74.4 & 71.3 & 75.0 & 75.2 & 75.2 & 75.8 & 76.0 & 74.3 & 75.1 \\
    MobileNetV2 \cite{mobileNetv2} & 68.8 & 66.5 & 68.2 & 72.6 & 72.8 & 69.7 & 72.4 & 69.5 & 73.1 & 72.9 & 72.7 & 73.6 & 73.4 & 72.4 & 72.9 \\ 
    ResNeXt50 \cite{resNext_2017}  & 71.7 & 69.4 & 71.0 & \textbf{76.9} & \textbf{77.6} & \textbf{75.8} & \textbf{76.5} & \textbf{75.8} & \textbf{77.4} & \textbf{77.4} & \textbf{76.9} & \textbf{77.8} & \textbf{78.1} & \textbf{76.0} & \textbf{76.2} \\ 
    Swin-B  \cite{swin_2021}    & 71.5 & 67.1 & 70.2 & 74.0 & 74.6 & 74.2 & 74.8 & 74.1 & 75.6 & 75.5 & 75.1 & 76.1 & 76.2 & 74.8 & 75.2 \\ 
    ConvNeXt-B \cite{convNext_2022} & \textbf{72.7} & \textbf{69.5} & \textbf{71.4} & 74.9 & 75.7 & 74.5 & 75.2 & 74.5 & 76.2 & 76.0 & 75.6 & 76.7 & 76.4 & 75.2 & 75.3 \\ 
    \noalign{\hrule height 1pt} \hline
    \end{tabular}
\end{table*}

\subsection*{Semantic Features}
Table \ref{tab:ssf_results} shows the accuracy values, regarding the semantic features-only evaluation, achieved on the SUN RGB-D \cite{sun_dataset} and NYU Depth v2 \cite{nyu_dataset} datasets. Promising results were obtained by the proposed SSFs on both SUN RGB-D and NYUv2 datasets, reaching 43.5\% and 67.3\% using the S50 segmentation model, respectively. Such performance obtained on the NYUv2 dataset is close to those achieved by the backbone networks on the RGB and segmentation mask data (results in Table \ref{tab:nyu_results}).
Overall, reported results show that extracting semantic features from segmentation masks generated by the S50 model significantly improved the performance when compared with the performance attained by the semantic features using the S20 model. The performance of the semantic features was again improved when the S75 model was used. Such performance behavior was expected, since segmentation models with high performance levels can generate segmentation masks representing well-defined category boundaries. In contrast, segmentation models
In contrast, segmentation models with lower performance levels may yield masks that inadequately represent the indoor scene, consequently degrading the quality of the extracted SSFs.
AP features achieved the lowest accuracy values, meaning that AP features had a minor influence in characterizing indoor scenes than the remaining SSFs-based features. 
On the SUN RGB-D dataset, the PC features attained the highest accuracy value compared with the AP features and SD features. On the NYUv2 dataset, PC and SD features obtained similar accuracy values. However, on both datasets, combining multiple SSFs-based features led to performance improvements, showing that combining the proposed features can improve the feature representation of the scene. Furthermore, reported results on the SUN RGB-D dataset show that exploiting SSFs features through convolutional layers has slightly outperformed the performance obtained by using fully-connected layers. On the NYUv2 dataset, exploiting the SSFs through convolutional or fully-connected layers presented similar performances.

\subsection*{Global and Semantic Feature Fusion}
Tables \ref{tab:sun_results} and \ref{tab:nyu_results} present the accuracy values obtained by the proposed approach and intermediate evaluations using different global feature extraction approaches as baseline/backbone on the SUN RGB-D and NYUv2 datasets, respectively. Overall, very promising results were achieved by the GS$^2$F$^2$App, significantly improving the baseline results, except for the MobileNetV2 backbone network on the SUN RGB-D dataset. Such results highlight the positive impact of the proposed approach in obtaining a discriminative feature representation of indoor scenes. On the SUN RGB-D dataset, the ConvNeXt-B backbone network achieved the highest performance (RGB), 58.3\%. Using it in the GS$^2$F$^2$App resulted in the highest accuracy value, 62.3\% (S50). A comparable accuracy value was attained using the ResNeXt50 backbone network, which itself achieved an RGB-only performance of 56.9\%.
Furthermore, comparing the results obtained by the backbone networks with those achieved by the GS$^2$F$^2$App, it can be observed that the proposed approach is not reliant on a specific backbone network. The overall performance of the proposed approach relies on the synergy between the representation of the global feature space and the representation of semantic features.
On the NYUv2 dataset, a similar trend emerges, with the GS$^2$F$^2$App achieving its highest accuracy of 77.8\% (S50) when using the ResNeXt50 backbone network. In contrast, the ConvNeXt-B backbone network achieves the highest RGB-only performance at 72.7\%. Reported results on both datasets also show that the segmentation model performance used to extract the SSFs, S50 or S75, does not present a significant influence on the overall GS$^2$F$^2$App performance. This may happen due to the dataset having a high number of scene categories with inter-category ambiguity challenges, as reported in \cite{gsf2app}, and the used segmentation-categories may not be enough to uniquely represent all scene categories. Conversely, it is interesting to note that the feature representation built using the S50 model, with the proposed SSFs combined with global features, is discriminative with respect to the scene classification categories and that improving the segmentation model’s performance produces similar results in scene classification performance.

The proposed approach achieved higher accuracy values when SSFs were exploited by the SSFs-CNN instead of FC layers, showing that using convolutional layers to exploit the SSFs led to a better scene representation of indoor scenes. Furthermore, on the SUN RGB-D dataset, the GS$^2$F$^2$App slightly outperformed the RGB+SegMask. In contrast, on the NYUv2 dataset, the proposed approach presents a significant improvement over the RGB+SegMask. Such results highlight the effectiveness of the proposed SSFs in encoding how segmentation-categories are distributed in the scene. Moreover, exploiting SSFs has a lower computational burden than applying a CNN to a segmentation mask, which can be an important aspect when dealing with limited computational resources (see Table \ref{tab:performance}). Regarding the intermediate evaluations of the SSFs-based features, combining global features with the AP features resulted in higher accuracy values when compared to the use of PC or SD features. 
These results underline the effectiveness of using AP features, which, when integrated with global features, provide a more fitting representation of indoor scenes.
Reported results also show a performance improvement when multiple SSFs-based features were combined, showing that a better feature representation of indoor scenes was achieved.

\setlength{\tabcolsep}{5.8pt}
\begin{table}[!tb]
    \centering
    \footnotesize
    \caption{Computational complexity.}
    \label{tab:performance}
    \begin{tabular}{lccc}
        \noalign{\hrule height 1pt} \hline
        Network                  & FLOPs (B) & \# Params (M) & FPS  \\ \hline \hline
        SSFs-CNN                 & 0.0812 & 12.62         & 1133 \\
        SSFs-NN                  & 0.0015 & 0.73          & 8748 \\ \hline
        RGB (Swin-B)             & 20.46  & 59.25         & 61   \\
        RGB (ResNeXt50)          & 8.57   & 24.04         & 174  \\
        RGB (ConvNeXt-B)         & 30.75  & 88.08         & 111  \\ \hline
        RGB+SegMask (ResNeXt50)  & 17.15  & 48.59         & 88   \\
        RGB+SegMask (ConvNeXt-B) & 61.39  & 176.67        & 54   \\ \hline
        GS$^2$F$^2$App (Swin-B)        & 20.53  & 72.64         & 59   \\
        GS$^2$F$^2$App (ResNeXt50)     & 8.66   & 37.43         & 162  \\
        GS$^2$F$^2$App (ConvNeXt-B)    & 30.83  & 101.47        & 105  \\ \hline
        GS$^2$F$^2$App-NN (ResNeXt50)  & 8.58   & 25.28         & 173  \\
        GS$^2$F$^2$App-NN (ConvNeXt-B) & 30.75  & 89.32         & 109  \\ \noalign{\hrule height 1pt} \hline
        \end{tabular}
\end{table}

\subsection*{GS$^2$F$^2$App Computational Complexity}

The computational complexity of the GS$^2$F$^2$App, presented in Table \ref{tab:performance}, was measured through the following metrics: Floating Point Operations (FLOPs), the network's number of parameters, and Frames Per Second (FPS). The SSFs-CNN exhibits a low FLOPs value compared to any FLOPs value obtained by the backbone networks. It also achieved a very high FPS value. Backbone networks also demonstrated high FPS values. The GS$^2$F$^2$App and the employed backbone network exhibit similar FLOPs and FPS values. This occurs due to the low FLOPs value and very high FPS value achieved by the SSFs-CNN. Consequently, the GS$^2$F$^2$App, whether using the ResNeXt50 or ConvNeXt-B backbone network, achieved high FPS values, running at 162 and 105 FPS, respectively. 

Overall, reported results demonstrate positive evidence of the GS$^2$F$^2$App’s applicability in real-time applications. The values presented in Table 5 do not consider the DeepLabv3+ network.

\section{Conclusion}

In this paper, aiming to obtain meaningful semantic information about an indoor scene that led to a more complete feature representation, a new segmentation-based approach to extract SSFs, was proposed. They encode a 2D-based spatial distribution of the segmentation-categories over the scene.
Moreover, the SSFs were integrated in a two-branch CNN architecture, GS$^2$F$^2$App, which also exploits CNN-based global features extracted from RGB images. 
The GS$^2$F$^2$App underwent evaluation on two benchmark indoor scene datasets,  SUN RGB-D and NYU Depth V2, achieving accuracy values of 62.3\% and 77.8\%, respectively. As far as we know, these performances represent the highest reported results in the literature, underscoring the significant positive impact of the proposed approach on the indoor scene classification task. Reported results showed that the GS$^2$F$^2$App significantly improves the backbone network performances, and it does not have a specific backbone network dependency. 
Furthermore, the GS$^2$F$^2$App's overall performance is not significantly impacted by whether the SSFs are extracted using the S50 or S75 segmentation model. Reported results also showed that the SSFs-based AP features seem to have a greater influence than the SSFs-based PC or SSF-based SD features in achieving a better feature representation of indoor scenes. However, despite the promising results achieved by the GS$^2$F$^2$App, the SSFs present some limitations. For instance, they do not encompass correlated features between segmentation-categories nor provide semantic information about the number of objects and their respective categories available in the scene.
In future work, seeking more meaningful semantic information about indoor scenes, merging semantic features extracted from semantic segmentation masks and semantic features extracted from object detection techniques can be exploited.

\section*{Acknowledgments}

Ricardo Pereira and Tiago Barros have been supported by the Portuguese Foundation for Science and Technology (FCT) under the PhD grants with references SFRH/BD/148779/2019 and 2021.06492.BD, respectively. This work has been also supported by the FCT through grant UIDP/00048/2020 and partially funded by Agenda “GreenAuto: Green innovation for the Automotive Industry”, with reference C644867037.


\bibliographystyle{IEEEtran}
\bibliography{PRL_Final_references}

\end{document}